\documentclass{article}

\PassOptionsToPackage{square}{natbib}
\usepackage[preprint]{neurips_2021}

\usepackage[utf8]{inputenc} % allow utf-8 input
\usepackage[T1]{fontenc}    % use 8-bit T1 fonts
\usepackage{hyperref}       % hyperlinks
\usepackage{url}            % simple URL typesetting
\usepackage{booktabs}       % professional-quality tables
\usepackage{amsfonts}       % blackboard math symbols
\usepackage{nicefrac}       % compact symbols for 1/2, etc.
\usepackage{microtype}      % microtypography
\usepackage{xcolor}         % colors
\usepackage{graphicx}
\usepackage{wrapfig}
\usepackage{varwidth}
\usepackage{enumitem}
\usepackage{authblk}
\usepackage{amsmath}
\usepackage{pdfpages}
\setcitestyle{numbers}
\hypersetup{
    colorlinks=false,
    pdfborder={0 0 0},
    pdfborderstyle={/S/U/W 0},
}

\DeclareMathOperator*{\argmax}{argmax}
\newtheorem{definition}{Definition}[section]
\begin{document}

\title{A Pareto-optimal compositional energy-based model for sampling and optimization of protein sequences}

\author[1]{\textbf{Nata\v{s}a Tagasovska}}
\author[1]{\textbf{Nathan C. Frey}}
\author[1]{\textbf{Andreas Loukas}}
\author[2]{\textbf{Isidro Hötzel}}
\author[2]{\textbf{Julien Lafrance-Vanasse}}
\author[2]{\textbf{Ryan Lewis Kelly}}
\author[2]{\textbf{Yan Wu}}
\author[2]{\textbf{Arvind Rajpal}}
\author[1]{\textbf{Richard Bonneau}}
\author[1,3,4,5]{\textbf{Kyunghyun Cho}}
\author[1]{\textbf{Stephen Ra}}
\author[1]{\textbf{Vladimir Gligorijevi\'c}}

\affil[1]{\footnotesize Prescient Design, Genentech}
\affil[2]{\footnotesize Antibody Engineering, Genentech}
\affil[3]{\footnotesize Department of Computer Science, Courant Institute of Mathematical Sciences, New York University}
\affil[4]{\footnotesize Center for Data Science, New York University}
\affil[5]{\footnotesize CIFAR Fellow}
\affil[ ]{\texttt{natasa.tagasovska@roche.com}}

\maketitle
\begin{abstract}
  Deep generative models have emerged as a popular machine learning-based approach for inverse design problems in the life sciences.
  However, these problems often require sampling new designs that satisfy multiple properties of interest in addition to learning the data distribution.
  This multi-objective optimization becomes more challenging when properties are independent or orthogonal to each other. 
  In this work, we propose a \emph{Pareto-compositional energy-based model (pcEBM)}, a framework that uses multiple gradient descent for sampling new designs that adhere to various constraints in optimizing distinct properties. We demonstrate its ability to learn non-convex Pareto fronts and generate sequences that simultaneously satisfy multiple desired properties across a series of real-world antibody design tasks.
\end{abstract}

\section{Introduction}
\label{sec:intro}

Generative models have shown promise across various applications in the life sciences for generating chemically- and physically-plausible designs and in accelerating the process of scientific discovery. 
Part of this trend of adoption can be owed to  the convincing examples created by generative adversarial networks (GANs) \cite{goodfellow2020generative}, variational autoencoders (VAEs) \cite{kingma2013auto}, energy-based models (EBMs) \cite{du2019implicit} and, more recently, diffusion models \cite{sohl2015deep, Roose2022, connard2022}. However, there are far fewer success stories in real-world industry applications \cite{david2020,Hao2022}. 
Some reasons include an overrepresenation of image datasets; a lack of evaluation protocols and metrics for synthetic data \cite{theis2015note, sajjadi2018assessing}; challenges around controllable generation and training --- for exmaple GANs; and challenges in generating samples that are different from have been seen during training \cite{karras2017progressive, barz2020we, van2021memorization}.
Taken together, these serve to limit the applicability of generative modeling for real-world use cases. 

Our work is motivated by this and, in particular, the need to accelerate the development and discovery of new molecules, namely therapeutic antibodies.
Though several generative models have already been proposed for these purposes~\cite{eguchi2022ig, gligorijevic2021function, repecka2021expanding}, generating new samples without guidance or control does not guarantee downstream success of the proposed molecules. 
In practice, each molecule has to satisfy multiple properties. For therapeutic antibodies, this could include properties such as binding affinity, polyreactivity, and viscosity~\cite{abdev_review,jain2017biophysical,wolf2022}. 
Failure to  account for these and other properties can lead to serious complications later on during scale-up, manufacturing, and clinical  trials and the optimization of straightforward objectives does not necessarily translate into progress in the laboratory~\cite{nigam2022tartarus}. 

Motivated by this challenge, we propose a new EBM for antibody design that simultaneously takes into account multiple properties that an antibody has to satisfy.
The adherence to multiple properties is a challenge in itself, as it often involves optimizing multiple conflicting objectives. 
In antibodies, for instance, optimization of binding affinity alone may come at the expense of developability properties, parameters that govern the likelihood of success of a molecule during manufacturing and quality control .  

\begin{figure}
    \centering
    \includegraphics[width=0.85\textwidth]{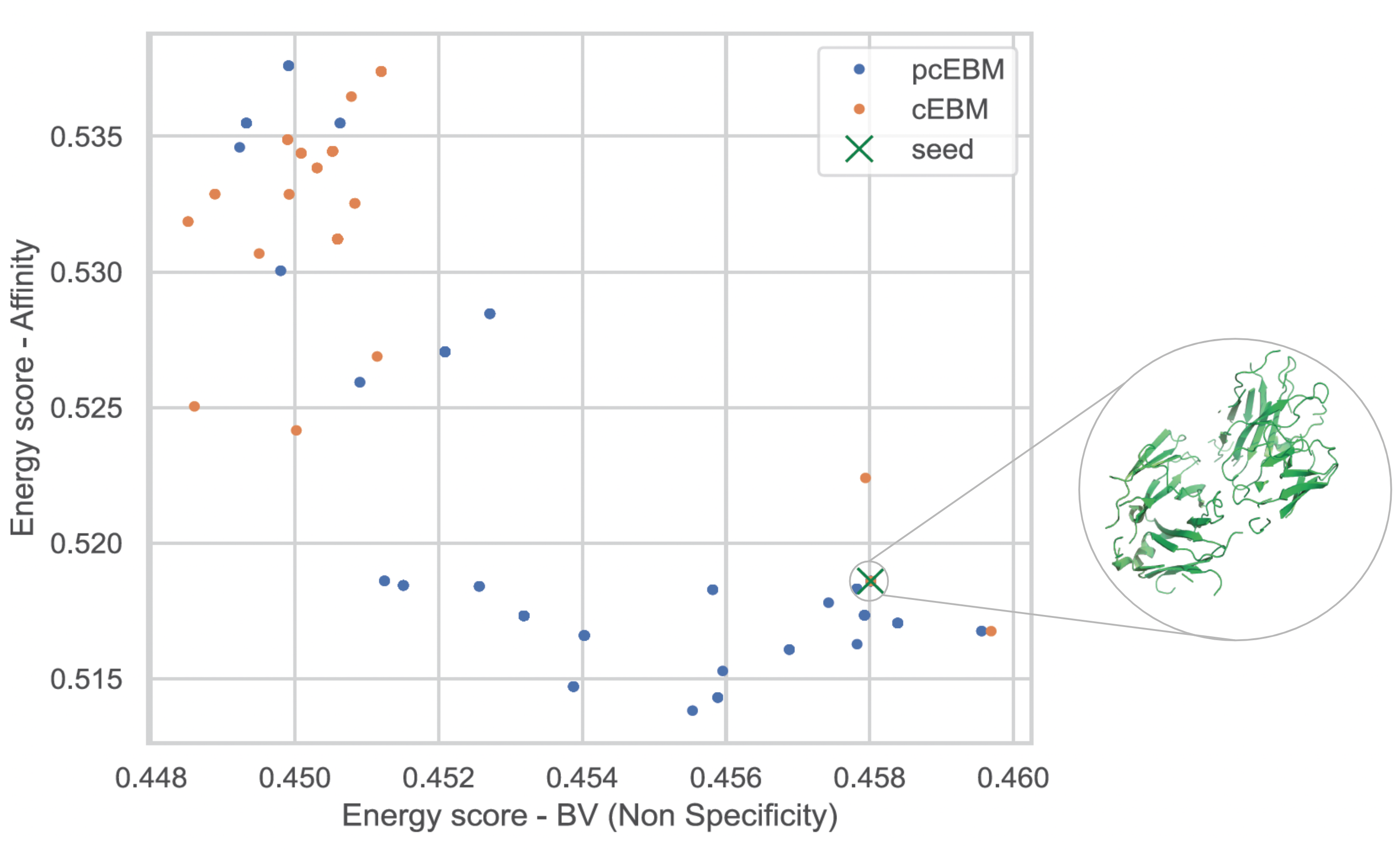}
    \caption{Example output of the proposed Pareto-optimal energy-based sampler (pcEBM) compared to a naive multi-objective sampling (cEBM). The green marker denotes the starting sequence; each point in the plot corresponds to a modified design of that starting point, aiming at improving affinity and nonspecificity (BV-ELISA or BV score). pcEBM introduces samples along a Pareto front, with candidates minimizing both objectives simultaneously, while candidates generated by the cEBM minimize only one objective.}
    \label{fig:intro}
\end{figure}

In general, in optimizing multiple conflicting properties, it is often impossible to find a single sample that satisfies all of the objectives simultaneously \cite{sayin2003kaisa}. 
We argue that, instead, one should aim to propose a set of diverse data points from the \emph{Pareto front} \cite{censor1977pareto} that correspond to different choices for various objective functions. Doing so enables a global perspective the optimal trade-off between objectives and can select a molecule according to their preference.
To achieve Pareto optimality, we rely on recent advances in multi-objective optimization (MOO)\cite{sener2018multi,liu2021profiling} as well as on compositional sampling with EBMs \cite{du2019implicit} to build a Pareto-optimal compositional EBM (\emph{pcEBM}). 
\autoref{fig:intro} exemplifies both why we need pcEBM and what we achieve with it when optimizing a design of an initial antibody sequence.

We first present the the necessary background on multi-objective optimization in \autoref{sec:mpo} and related work on compositional sampling with EBM in \autoref {sec:ebm} leading to pcEBM described in \autoref{sec:pcebm}. 
In \autoref{sec:experiments} we include empirical evaluation and discussion, before concluding in \autoref{sec:conclusion}.

\section{Background and method}  
\label{sec:background}

\vspace{2mm} \textbf{Problem setup.} 
Antibodies are composed of two chains of amino acids (AA), which can be represented as sequences of characters. Each AA comes from an alphabet of 20 characters corresponding , typically of combined length $L \sim 250$. 
We will denote the sequences as $\mathbf{}{x} = (x_1, \ldots, x_L)$, where $x_l \in \{1, \ldots, 20\}$ 
corresponds to the AA type at position $l$. 
For each of those sequences, we measure $m$ properties, such that $f_i: \mathbb{R}^L \rightarrow \mathbb{R}$ for all $i \in 1, \ldots, m$. 

Our goal is to generate new sequences $x^*$ with preferred values for each of the $m$ properties. 
Since we cannot fully satisfy all of the properties/objectives simultaneously (they may be conflicting with each other), we are interested in finding samples which can not be further improved simultaneously for all the objectives, yielding the notion of \emph{Pareto optimality}~\cite{chinchuluun2007survey}.

\begin{definition}[Pareto Optimality \cite{chinchuluun2007survey, liu2021profiling}]
For $x_1, x_2 \in \mathbb{R}^d$, we say that $x_1$ is \emph{dominated} by $x_2$ iff $f_i(x_2) \leq f_i(x_1), \forall i \in [m]$. 
A point $x^*$ is called \emph{globally Pareto optimal} on $\mathbb{R}^d$ iff it is not dominated by any other $x^\prime \in \mathbb{R}^d$. 
A point $x*$ is called \emph{locally Pareto optimal} iff there exists an open neighborhood $\mathcal{N}(x^*)$ of $x^*$, such that $x^*$ is not dominated by any $x \in \mathcal{N}(x^*)$. %
The collection of globally (resp. locally) Pareto optimal points are called the global (resp. local) Pareto set.
The collection of function values $F(x^*) = (f_1(x^*), f_2(x^*), \dots , f_m(x^*))$ of all the Pareto points $x^*$ is called the \emph{Pareto front}.
\end{definition}

\subsection{Multi-objective optimization}
\label{sec:mpo}

A simple approach to solving MOO is \emph{linear scalarization} \cite{sayin2003kaisa} based on a preference vector $\lambda = [\lambda_1, \ldots, \lambda_m]$ from the probability simplex on $[m]$, i.e., $S =  \lbrace \lambda: \sum_{i=1}^m \lambda_i = 1, \lambda_i \geq 0, i \in [m] \rbrace$.
Each $\lambda \in S$ leads to a weighted objective function $f_\lambda(x) = \sum_{i=1}^m \lambda_i f_i(x)$ and its minimizer $x^*_\lambda = \text{argmin}_x f_\lambda(x)$. 
As we evaluate on $\lambda$ from a grid of $S$, we hope that the corresponding $x^*_\lambda$ approximates the Pareto front. 
The solutions obtained by linear scalarization lie in the convex envelope of the Pareto front, which renders the approach plausible for situations where the Pareto front is convex \cite{liu2021profiling}. 

An alternative approach for non-convex Pareto fronts is \emph{multiple gradient descent (MGD)} \cite{desideri2012multiple}, which iteratively updates variable $x$ along a direction that ensures that all objectives are decreased simultaneously and guides the optimization towards a Pareto improvement.
Denote the gradient of the $i$-th objective as $\nabla_x f_i(x)$. 
With gradient descent, we update the variable $x^k \leftarrow x^{k-1} -\eta g(x)$, where $g(x)$ is a vector to be determined and $\eta$ is a small step size. From a 1st-order Taylor expansion, we can deduce that $\langle \nabla_x f_i(x), g \rangle \approx -(f_i(x^k) - f_i(x^{k-1}))/\eta$ which represents the decreasing rate of $f_i$ when we update $x$ along direction $g(x)$.
In MGD, $g$ is chosen to maximize the slowest decreasing rate among all the objectives, that is:
 \begin{align}
      g(x) \propto \argmax_{g \in \mathbb{R}^d} \lbrace  \min_{i \in [m]} \langle g, \nabla_x f_i(x) \rangle \ \text{subject to}  \ \|g\|_2 \leq 1\rbrace
 \end{align}
By doing so, $g(x)$ is pushed to have positive inner products with all $\nabla_x f_i(x)$. 
If the latter is impossible, $\lbrace \nabla_x f_i(x)\rbrace_{i=1}^m$ will contain conflicting directions and we will get $g(x) = 0$ which terminates the procedure.
By using Lagrangian duality, D\'esid\'eri \cite{desideri2012multiple} has shown that the above objective has the optimal value $g(x) \approx \sum_{i=1}^m \lambda_i^*\nabla_x f_i(x)$  where $\lbrace \lambda_i \nabla_x f_i(x)\rbrace_{i=1^m}$ is the solution of
$$
\min_{\lambda_i} \| \sum_{i=1}^m \lambda_i \, \nabla_x f_i(x) \|_2  \ \text{subject to}  \ \sum_{i=1}^m \lambda_i = 1  \ \text{and} \ \lambda_i \geq 0  \ \text{for all}  \ i \in [m].
$$
The above convex optimisation problem has a closed form solution when $m=2$ and a fast algorithm for $m > 2$ \cite{sener2018multi, jaggi2013revisiting}. %\andreas{should we write it down at least in the appendix?} 
By construction, when the step size is small, MGD decreases monotonically all objectives simultaneously and will terminate at a local Pareto point.

\subsection{Compositional energy-based models}
\label{sec:ebm}

Although EBMs have been around for many years \cite{lecun2007energy, lecun2006tutorial}, recent interest in generative modeling has led to a resurgence of interest \cite{du2019implicit, du2020compositional}.
EBMs learn to represent data by approximating an unnormalized probability distribution across data. 
They do so by learning an energy function $E_\theta(x)$ parameterized by a neural network that maps each input $x$ to a scalar real value interpreted as the energy and approximating the data distribution by the Boltzmann distribution under unit temperature: 
\begin{align*}
    p_\theta(x) \propto e^{-E_\theta(x)}.
\end{align*}
Training an EBM on a given data distribution is usually done by contrastive divergence \cite{hinton2002training}.
Besides predicting the un-normalized data likelihood, EBMs can be used to sample new data points $x$ from $p_\theta$ (during both training and generation).
Sampling is typically performed by some Markov-Chain Monte-Carlo (MCMC) procedure. 
We consider the common case of Langevin Diffusion (LD) where the samples are initialized from refinement/uniform random noise followed by iterative refinement:
\begin{align}
    x = x^{k-1} - \frac{\eta}{2}\nabla_{x}E_{\theta}(x^{k-1}) + \omega^k, \ \omega \sim \mathcal{N}(0, \sigma^2),
\end{align}
with $k$ being the sampling step and $\eta$ the step size. 

The above sampling procedure can be easily extended to produce samples from the composition of $p_\theta$ with other distributions of interest.
Along this direction, Du et al.~\cite{du2020compositional} considered the problem of sampling from $m$ composed distributions, each modeling a different propertiey $f_i$, an approach referred to as compositional EBM (cEBM). 
Du et al. proposed multiple variations for executing different logical operations over properties of interest, such as conjunction, negation, and disjunction. 
In the following, we focus on conjunction, which corresponds to combining individual EBMs as a product of experts:
$$
p(x|f_1 \wedge f_2 \wedge \cdots \wedge f_m) = \prod_i p(x|f_i) \propto e^{-\sum_i E(x|f_i)}.
$$
We also note that cEBM utilized the following sampling procedure:
\begin{align}
    x = x^{k-1} - \frac{\eta}{2}\nabla_{x} \sum_i E_{\theta}(x^{k-1} | f_i) + \omega^k, \  \omega \sim \mathcal{N}(0,\sigma^2),
\end{align}
\label{eq:cebm_sampl}
which amounts to applying LD w.r.t. $E(x) = \sum_i E(x|f_i)$. 

\subsection{Pareto compositional sampling}
\label{sec:pcebm}

Inspired by previous work on multi-objective optimization~\cite{desideri2012multiple,liu2021profiling} discussed in \autoref{sec:mpo}, we here propose pcEBM, a method that samples from the Pareto front of $m$ Boltzmann distributions of interest. 
Our approach entails utilizing multiple gradient descent to select a locally optimal direction that optimizes the slowest decreasing rate among all the objectives of the compositional EBMs:  
\begin{align}
    x^k \leftarrow x^{k-1} - \eta \argmax_{g \in \mathbb{R}^d} \lbrace  \min_{i \in [m]} \langle g, \nabla_x f_i(x^{k-1}) \rangle \ \text{subject to}  \ \|g\|_2 \leq 1\rbrace + \sqrt 2\alpha\omega,
    \label{eq:pcebm}
\end{align}
where $\alpha$ is a positive constant and the internal optimization problem is the same utilized in MGD and the external sampler is performing Langevin diffusion. 

% \andreas{The following needs re-writing imo. 1) It reads like a related work section whereas it should be a contribution. 2) The equation (4) has already been explained so it reads as somewhat repetitive. 3) The pcEBM is not clearly explained. Why not just write down what we do and then expose the connections to previous work after?}

% Liu et al.~\cite{liu2021profiling} improved MGD multi-objective optimization via sampling, encouraged by the opportunity to access a diverse set of solutions i.e., Pareto front (not only a single point local Pareto estimate), and importantly, compared to existing methods, without relaying on predefined preference functions (as in Linear scaling). The main idea is to incorporate the MGD minimum norm gradient updates within sampling optimization procedures as Stein variational gradient descent \cite{liu2017stein} or Langevin dynamics \cite{welling2011bayesian}: 
% % 
% \begin{align}
%     x^k \leftarrow x^{k-1} - \eta g^*(x) + \sqrt 2\alpha\omega, \ \omega \sim \mathcal{N}(0,\sigma^2),
% \end{align}
% \label{eq:mgd_ebm}
% % 
% where $\alpha$ is positive constant scaling the Gaussian noise $\omega$. 

% Recalling that $g^*(x) \approx \sum_{i=1}^m \lambda_i^*g_i(x)$ and that $g_i (x) = \nabla_i f_i(x)$, and replacing the gradients in \autoref{eq:cebm_sampl} with minimum norm gradients
% \andreas{what is a minimum norm gradient?}, 
% leads us to the Pareto optimal compositional EBM (\emph{pcEBM}). 

 \begin{wrapfigure}[20]{r}{0.45\textwidth}
    \vspace{-0.5cm}
    \includegraphics[width=0.45\textwidth]{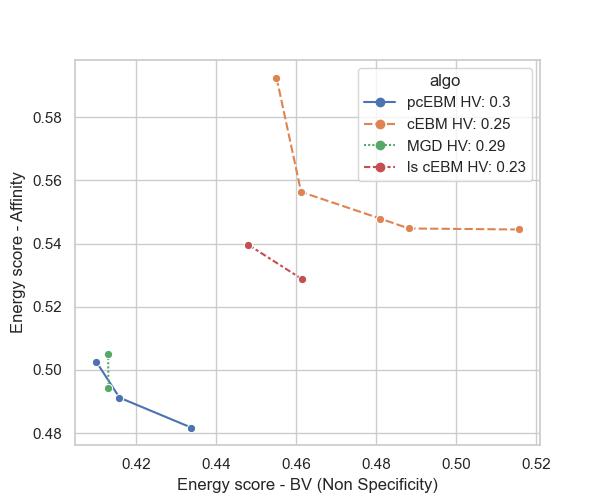}
    \caption{Pareto fronts and estimated hyper volume for all baselines with $\eta=0.01$ and $k=400$. The axes show the energy of EBMs that were separately trained on data with good affinity and BV scores, respectively, with smaller energies being more desirable.}
    \label{fig:pareto_fronts}
\end{wrapfigure}

Suppose that we are generating a new sequence $x$, either by starting from random noise or by improving an existing one. 
At each step $k$, if $x^{k-1}$ is far from the Pareto front and the gradients $\nabla_x f_i(x^{k-1})$ have non-negligible norm, \autoref{eq:pcebm}  will drive $x^{k-1}$ closer to the Pareto front.
On the other hand, when $x^*$ is close to the Pareto front and the gradients have nearly vanished, the noise dominates and $x^{k-1}$ performs Brownian motion. 
As will be demonstrated experimentally, the addition of noise $\omega$ helps pcEBM to explore more efficiently the Pareto as compared to MGD, whereas the locally optimal determination of the optimization direction renders the optimization more effective than cEBM. 

% Differently from \cite{liu2021profiling} that uses MGD with sampling for multi-property optimization, here, we leverage the joint framework of MGD and cEBM for multi-property sampling.

\section{Empirical evaluation}
\label{sec:experiments}
In the following, we compare the Pareto compositional EBM \emph{pcEBM} with three baselines: multiple gradient descent (\emph{MGD})~\cite{desideri2012multiple}, a compositional EBM (\emph{cEBM})~\cite{du2020compositional}, and a linearly scaled cEBM (\emph{ls-cEBM})~\cite{sayin2003kaisa}.

\textbf{Experimental setup.} 
We wish to generate plausible antibody sequences according to three key properties: (i) similarity to known human antibodies from the public database of observed antibodies~\cite{olsen2022observed} (\emph{Ab-like}), (ii) binding affinity to antigens of interest (\textit{Aff}), and (iii) an experimental measure of nonspecificity --- phenomena which can affect the dosing regimen, safety, and overall therapeutic profile of an antibody --- which we refer to as the \emph{BV score} ~\cite{hotzel2012strategy} \footnote{An enzyme-linked immunosorbent assay (ELISA) of non-specific binding to baculovirus particles ~\cite{hotzel2012strategy}.}. 
We use datasets from both public and proprietary sources to train three separate EBMs, each pertaining to a property of interest.
Though alternative choices are also possible, we instantiate all models as sequence-based convolution networks \cite{gehring2017convolutional}.
More details on the architecture can be found in the Appendix. 
We use the same models for sampling new sequences with each of the baselines.
The differences in implementation are as follows: \emph{MGD} does not add any noise, \emph{cEBM} uses Lanvegin dynamics but has uniform weights for all properties; \emph{ls-cEBM} differs from cEBM in that it includes domain-informed weights; and finally \emph{pcEBM} learns the optimal weights for each property per sequence.
Each baseline relies on two main hyper-parameters, namely the \emph{step size} $\eta$ and the \emph{number of steps} $k$.

To compare different approaches in terms of their ability to satisfy multiple properties, we adopt the Hyper-Volume (\emph{HV}) metric from the multi-objective optimization literature~\cite{guerreiro2020hypervolume} computed over the energy scores of each property model.
In all experiments, we aim for minimizing the energy scores per property.
Additionally, as a proxy for how well the sampled sequences capture the properties of interest we compute the minimal edit distance of each sampled sequence to the ones seen during the training of each property EBM (\emph{$e_{dist}$}) \cite{paassen2018tree}.
More details about the computation of these metrics can be found in the Appendix.

\vspace{2mm}\textbf{Results and discussion.} Our experiments are designed to provide answers to the following research questions:
\vspace{-1mm}
\begin{itemize}[noitemsep,topsep=0pt,leftmargin=.15in]
    \item \textbf{Q.1} Can we use EBMs to generate valid/reasonable, multi-property compliant antibody sequences? % (table edits dist and mmd/ood scores)
    \item \textbf{Q.2} Does sampling with Langevin Dynamics propose more Pareto optimal sequences? % (hypervolume table + graph)
    \item \textbf{Q.3} Does pcEBM have an advantage over cEBM?
    \item \textbf{Q.4} Starting from a seed sequence, can we improve a  property of interest using (p)cEBM? % (histograms of energies; bv elisa table scores; tsne plot)
\end{itemize}

\begin{table}[h]
\small
\centering
\resizebox{0.9\textwidth}{!}{
\begin{tabular}{lcccc}
\toprule
\textbf{} & \textbf{HV(Ab-Like, Aff, BV)}    & \textbf{HV(Aff, BV) } & \textbf{HV(Ab-Like, BV) } & \textbf{HV(Ab-Like, Aff) } \\ \midrule
$\eta = 0.01$ \\ \midrule           
\textbf{MGD}     & 0.04     & 0.22 & 0.23 & 0.29 \\
\textbf{ls-cEBM}     & 0.00     & 0.24 & 0.24 & 0.23 \\
\textbf{cEBM}     & 0.00     & 0.25 & 0.24 & 0.25 \\
\textbf{pcEBM}                         & \textbf{0.049} & \textbf{0.3}    & \textbf{0.29}   & \textbf{0.30}    \\ \midrule
$\eta = 1$ \\ \midrule               \\
\textbf{MGD}     & 0.04     & 0.25 & 0.25 & 0.27 \\
\textbf{ls-cEBM}     & 0.00     & 0.25 & 0.24 & 0.25 \\
\textbf{cEBM}     & 0.046 & 0.29 & 0.29 & 0.3  \\
\textbf{pcEBM} & 0.041 & 0.27 & 0.27 & 0.29 \\ \midrule
$\eta = 40$ \\ \midrule         \\
\textbf{MGD}     & 0.033     & 0.25 & 0.25 & 0.26 \\
\textbf{ls-cEBM}     & 0.043     & 0.27 & 0.26 & 0.26 \\
\textbf{cEBM}     & 0.046 & 0.29 & 0.29 & 0.29 \\
\textbf{pcEBM} & 0.037 & 0.27 & 0.27 & 0.28 \\ \bottomrule \\
\end{tabular}
}
\caption{Hyper-Volume (HV) across multiple combinations of properties (as measured by energy score) across different step sizes $\eta$. Larger scores indicate bigger improvement. The best results for are typeset in boldface.}
   \label{tab:hv_table}
\end{table}

\begin{table}[ht]
\centering
\small
\resizebox{0.9\textwidth}{!}{
\begin{tabular}{@{}lcccc@{}}
\toprule
\textbf{}                   & {$e_{dist}$(generated, Ab-like)} & {$e_{dist}$(generated, BV)} & {$e_{dist}$(generated, Aff)} & average $e_{dist}$\\ \midrule
single-property EBM \\ \midrule
\textbf{Ab-Like EBM}     &   98.74 ($\pm$19.25)    &   93.84 ($\pm$19.05)    & 103.3 ($\pm$17.8)  & 98.62   \\
\textbf{Stc EBM}     &  84.68 ($\pm$15.78)      &  \textbf{69.7 ($\pm$ 16.42)}     &   88.29 ($\pm$ 5.72) & 121.33   \\
\textbf{Aff EBM}     &    88.51 ($\pm$16.25)    &    81.4 ($\pm$15.6)   &   86.73 ($\pm$6.31)  & 85.55  \\ \midrule

$\eta = 40$ \\ \midrule
\textbf{MGD}     &  83.51 ($\pm$16.1)     & 71.49 ($\pm$15.7) & 85.84 ($\pm$16.3) & 80.28  \\
\textbf{cEBM}     & 92.96 ($\pm$ 19.03)       &  86.86 ($\pm$ 17.99) & 97.97 ($\pm$ 17.28) & 92.60    \\
\textbf{ls-cEBM}     & 92.4 ($\pm$ 19.3)      &  87.14 ($\pm$ 17.94) & 97.93 ($\pm$ 17.25) & 92.49    \\
\textbf{pcEBM} & \textbf{82.24 ($\pm$16.16)} & 71.47 ($\pm$ 15.82)     & \textbf{85.76 ($\pm$ 16.34)} & \textbf{79.82}     \\  \midrule
$\eta = 0.01$ \\ \midrule
\textbf{MGD}     & 91.46 ($\pm$16.23)    & 84.02 ($\pm$15.43)  &   92.91 ($\pm$16.48)  & 89.46   \\
\textbf{cEBM} & 219.23 ($\pm$4.43)   &  222.47 ($\pm$ 4.45)     &   221.99 ($\pm$ 4.4) & 221.23 \\
\textbf{ls-cEBM} & 218.78 ($\pm$4.66)    & 218.13 ($\pm$4.58)  &   218.62 ($\pm$4.73) & 218.51  \\
\textbf{pcEBM} & 89.35 ($\pm$ 14.64)     & 79.67 ($\pm$ 13.99)    & 89.04 ($\pm$ 14.37) & 86.02 \\     \bottomrule \\
\end{tabular}
}
\caption{Edit distance (mean with standard deviation in brackets) to sequences in the training set with property of interest. Results in bold are best per columns, smaller is better, considering $e_{dist}$ as a proxy for similarity to real data with the specific property.\label{tab:edists}}
\end{table}

\autoref{tab:hv_table} and \autoref{tab:edists} address \textbf{Q.1}. We find that EBMs trained on data possessing a single property of interest propose sequences that are similar to those sharing the same property, but are dissimilar to the sequences preferred by other objectives.
In contrast, sequences generated by a multi-property EBMs have smaller distances to all properties compared to the single-property EBMs. 
Amongst the four considered multi-objective baselines, pcEBM generates sequences with greater hyper-volume and similarity to the training datasets, whereas MGD is a close second. 

\begin{wrapfigure}[17]{r}{0.45\textwidth}
%\vspace{-0.1cm}
\centering
    \includegraphics[width=0.45\textwidth]{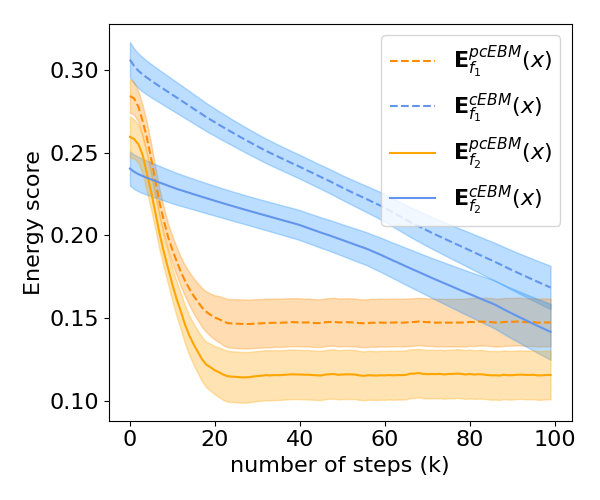}
    \caption{Energy scores per property $f_i$ at each step for pcEBM and cEBM; same sequence and same step size $\eta=0.01$.}
    \label{fig:pcebm_conv}
% \vspace{-0.6cm}
\end{wrapfigure}

We surmise the difference between cEBM and pcEBM is owed to their speed of convergence. We confirm this in \autoref{fig:pcebm_conv} by plotting the energy score at each sampling step for both cEBM and pcEBM with a step size of $\eta=0.01$. 
As observed, for the same starting sequence pcEBM converges faster. 
This result partially answers \textbf{Q.3}. 

To further test our approach, we calculate HV across the three properties we aim to account for.
Here, we consider the energy of each property EBM as proxy for the objective of interest\footnote{Though well suited to study the effect of multi-objective sampling, the energy score is biased since cEBM and ls-cEBM have the advantage of sampling and evaluating with respect to the same model.
A fairer comparison would be using external predictors as surrogates or wet-lab results to objectively evaluate.}. 
The advantage of pcEBM is noticeable at smaller step sizes (see \autoref{tab:hv_table}), while the variance in both edit distance and hyper-volume for different $\eta$ is significantly smaller for pcEBM than others. 
\autoref{fig:pareto_fronts} depicts the Pareto front for two properties obtained from candidate sequences from each baseline. We find that the pcEBM front is broader than MGD-based approaches, which agres our supposition that employing LD leads to an improved coverage of the Pareto front as compared to pure gradient descent. 
Further, pcEBM and MGD fronts are closer to the origin that that obtained by direct LD, which demonstrates the advantage of the multi-properly optimization aspect of pcEBM, addressing \textbf{Q.2} and \textbf{Q.3}.

\begin{figure}[t!]
    \centering
    {{\includegraphics[width=5cm]{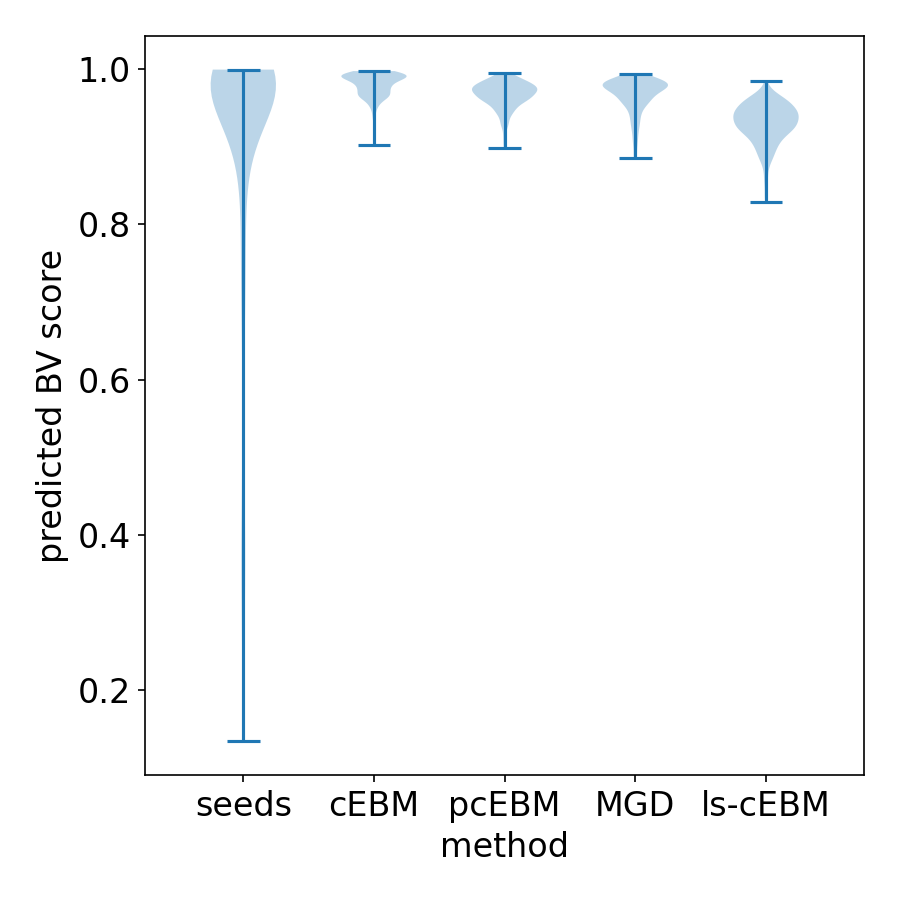} }}%
    \qquad
    {{\includegraphics[width=4.9cm]{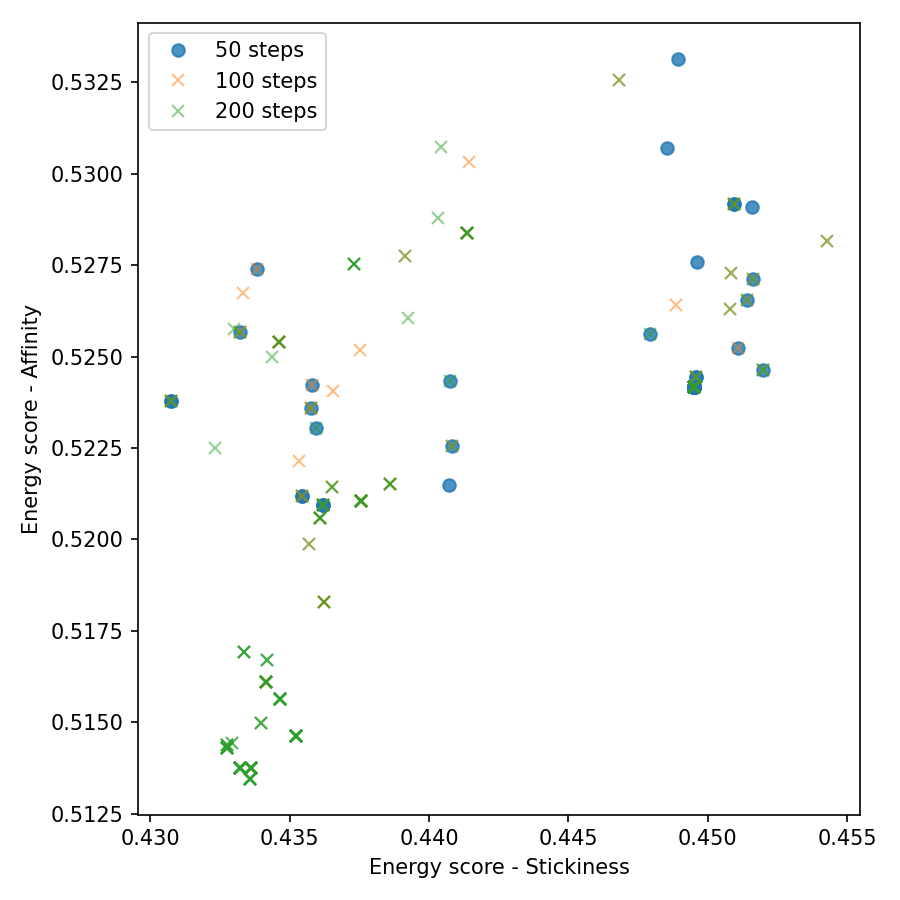} }}%
    \caption{ Left - violin plots of the distributions of predicted BV scores for the initial set of sequences to be improved (seeds), and for the proposed designs from each baseline. Higher BV scores are better. Right - energy scores of trajectories proposed by pcEBM for a different number of steps. Lower energy scores are better/preferred for the objectives.}%
    \label{fig:bv_violin}%
\end{figure}

Finally, for \textbf{Q.4} we wish to explore the possibility for improving properties of already existing antibodies by using the variants of compositional EBMs. 
We focus our analysis on nonspecificity (\emph{BV}) as we have an external SeqCNN classifier that can act as a surrogate pseudo-oracle. 
We start by screening the existing sequences to select only those with low BV scores, which we refer to as ``seeds''. 
We use sequences with poor experimental and predicted BV scores as seed sequences for generation and then reuse the nonspecificity pseudo-oracle to evaluate the newly proposed designs.
\autoref{fig:bv_violin} shows the BV score improvement achieved by each baseline. We notice that the multi-objective methods manage to improve the worst BV scores from 0.27 to above 0.9, with MGD and pcEBM giving a slightly lower average score than cEBM (though all scores are comparably high).  

\begin{figure}[t!]
    \centering
    {{\includegraphics[width=5.5cm]{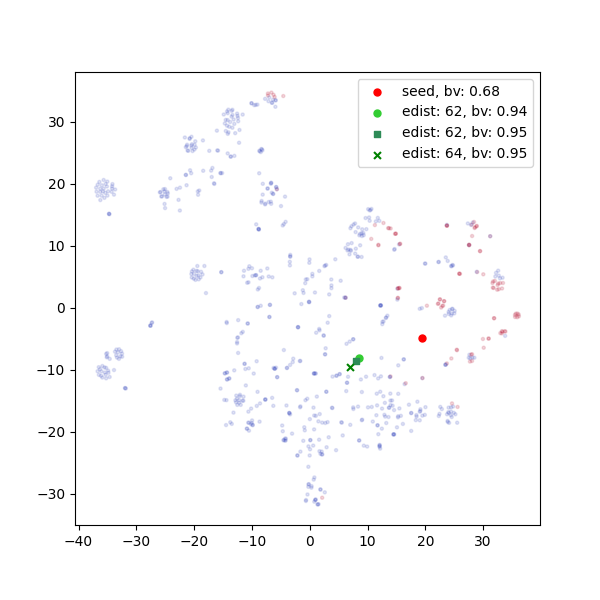} }}
    \qquad
    {{\includegraphics[width=5.5cm]{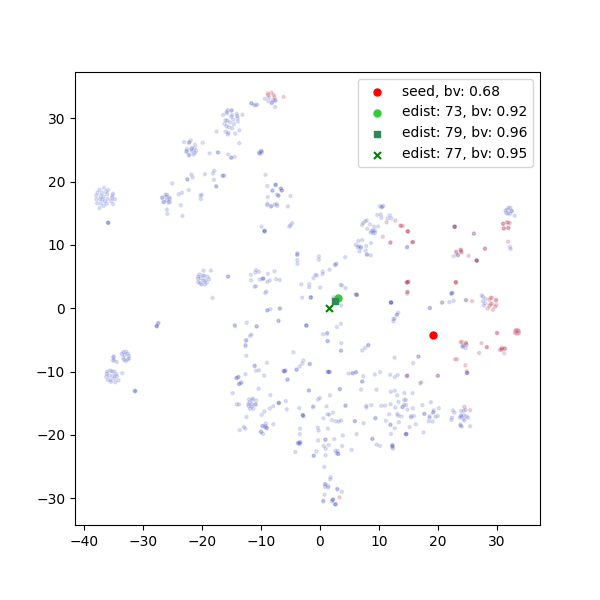} }}
    \caption{tSNE embeddings of antibody sequences colored based on BV score, with blue sequences having better scores. The figures show how a sequence with poor BV score of 0.68 (in orange) is improved by two models. Left - cEBM, Right - pcEBM.}
    \label{fig:tsne_mgdlms}
\end{figure}

%\andreas{Fig 4: add colorbar to indicate stickiness. Make the markers of designs larger and more visible/ (For camera-ready version: show the path during sampling).}
The plots in Figure~\ref{fig:tsne_mgdlms} provide further insight into the nonspecificity optimization process by examining a low-dimensional embedding of seed and optimized sequences. 
The embedding is obtained by using tSNE to project the last layer feature representations of the sequences obtained from the nonspecificity oracle, onto a 2D space. 
We color each point (sequence) by their BV score with blue indicating a good score and red a poor one. 
On the same plot, we overlay a seed sequence and its corresponding proposed designs.
We see that both cEBM and pcEBM improve the probability for a good nonspecificity score, and move the seed towards the section of the manifold covered by datapoints with good scores. 

\section{Conclusion}
\label{sec:conclusion}

We propose a new approach to sampling data points that simultaneously satisfy multiple desired properties. 
We do so by integrating multiple gradients within compositional energy based models.
In a series of experiments for generating and improving existing sequences, we validate the performance of the compositional EBMs as well as their Pareto optimality. 
The generality and modularity of pcEBM allow for multiple expansions we are excited to explore: (\textit{i}) other data types, in particular graph structures suitable for molecules, (\textit{ii}) incorporating other sampling frameworks such as Stein variational gradient descent \cite{liu2017stein}, (\textit{iii}) incorporating uncertainty estimates, and (\textit{iv}) Pareto-optimal training for EBMs, to name a few.

\bibliographystyle{abbrvnat}
\bibliography{mgdlms.bib}
%%%%%%%%%%%%%%%%%%%%%%%%%%%%%%%%%%%%%%%%%%%%%%%%%%%%%%%%%%%%

\newpage
\section*{Appendix}

\subsection{Additional details on experiments}

\paragraph{SeqCNNs} For all properties and baselines we use an identical NN architecture, that is one model per protein chain, consisted of three Conv1D layers with kernel size 9 and padding 1, ReLU non-linearities and penultimate layer of size 256. All EBMs were trained with contrastive training, Adam optimizer and early stopping criterion. 
\paragraph{Hyperparameters} For each multipropery sampling baseline, the following hyper parameter grid was explored with random search:
\begin{itemize}
    \item step size - $\eta \in \lbrace 1^{e-4}, 1^{e-2}, 1, 10, 40\rbrace$
    \item number of steps - $k \in \lbrace 100, 200, 300, 400\rbrace$
    \item noise type - $\omega \in \lbrace \mathcal{N}, \mathcal{U}\rbrace$.
\end{itemize}

\subsection{Metrics}

\paragraph{Hypervolume}

HV is an indicator for evaluating the quality of sets of solutions in multi objective optimization. For a given reference point of $r = [r_1, ..., r_m]$ we have an upper bound of
the objectives, such that $supx f_i(x) \leq r_i,\forall i \in [m]$. For a given set of solutions $\mathcal{X} = \lbrace x_l \rbrace^n_{l=1}$, a hypervolume indicator $HV(\mathcal{X})$ is a measure of the region between all $f_i$ objectives  and $r$:
\begin{align*}
    HV(\mathcal{X}) = \Lambda \{ (q \in \mathbb{R}^d \vert \exists x \in \mathcal{X}: q \in \prod_{i}^m [f_i(x), r_i] \} )
\end{align*}
where $\Lambda(\cdot)$ denotes the Lebesgue measure.

In the context of our experiments, we are interested in minimizing the energy scores per each property, and we choose a reference point $r = (1.0, 1.0, 1.0)$. We compte HV from each Pareto front to the reference point with the implementation provided in \verb|PyMoo| \cite{pymoo}.

\paragraph{Edit distance}
Edit distance is the typical score for quantifying how dissimilar two strings (in our case, sequences) are to one another, measured by counting the minimum number of operations required to transform one string into the other. Although different variants exist, here we rely on the  Levenshtein distance which accounts for insertions, deletions and substitutions. In all experiments we use the python library \verb|edist| as implemented in \cite{vsovsic2017edlib}.

\end{document}